\documentclass[letterpaper, 10 pt, conference]{ieeeconf}  

\IEEEoverridecommandlockouts                              

\usepackage{tikz}
\usepackage{comment}
\usepackage{lipsum}
\usepackage[normalem]{ulem}
\usetikzlibrary{positioning}
\usepackage{hyperref}
\usepackage{wrapfig}
\def\imagepath{./ascii}
\usepackage[imagepath=\imagepath]{ditaa}
\graphicspath{ {\imagepath/} }

\usetikzlibrary{%
  arrows,%
  shapes.misc,
  shapes.arrows,%
  chains,%
  matrix,%
  positioning,
  scopes,%
  decorations.pathmorphing,
  shadows%
}
\usepackage{chronosys}
\usepackage{placeins}
\catcode`\@=11
\def\chron@selectmonth#1{\ifcase#1\or January\or February\or March\or April\or%
 May\or June\or July\or August\or September\or October\or November\or December\fi}
\usepackage{pgf-umlsd}

\usepackage{colortbl}
\usepackage{pifont}
\usetikzlibrary{arrows,automata,positioning,calc}
\usepackage{wrapfig}

\newcommand\todo[1]{\textcolor{red}{#1}}

\usepackage{pgf-pie}
\usepackage{placeins}
\usepackage{pgfplots}
\pgfplotsset{compat = newest}
\usepackage{dblfloatfix}    

\usepackage{colortbl}
\usepackage{xcolor}
\usepackage{circledsteps}

\newcommand*\circled[1]{\tikz[baseline=(char.base)]{
    \node[shape=circle, draw, inner sep=1pt, fill=black, text=white, 
        minimum height=12pt] (char) {#1};}}

\newcommand*\circledwhite[1]{\tikz[baseline=(char.base)]{
            \node[shape=circle,draw,inner sep=1pt, minimum height=12pt] (char) {#1};}}

\usepackage{booktabs}
\usepackage{multirow}

\usepackage{flushend}

\title{\LARGE \bf RobotCore: An Open Architecture for Hardware Acceleration in ROS~2}

\author{Víctor Mayoral-Vilches$^{1}$$^{,2}$$^{,3}$, Sabrina M. Neuman$^{4}$,  Brian Plancher$^{4}$$^{,5}$,  Vijay Janapa Reddi$^{4}$
\thanks{This material is based upon work funded by Xilinx and supported by the National Science Foundation under Grant 2030859 to the Computing Research Association for the CIFellows Project. Any opinions, findings, conclusions, or recommendations expressed in this material are those of the authors and may not reflect those of the funding organizations.}%
\thanks{$^{1}$Víctor Mayoral-Vilches is with Acceleration Robotics,
        Ecuador 3, 1 I, Vitoria, Álava, Spain
        {\tt\small victor@accelerationrobotics.com}}%
\thanks{$^{2}$Víctor Mayoral-Vilches is with the System Security Group, Universit\"at Klagenfurt, Universitätsstr. 65-67 9020 Klagenfurt, Austria
        {\tt\small v1mayoralv@edu.aau.at}}%
\thanks{$^{3}$Víctor Mayoral-Vilches is with Alias Robotics,
        Venta de la Estrella 6, pab 130, Vitoria 01006, Spain
        {\tt\small victor@aliasrobotics.com}}%
\thanks{$^{4}$Sabrina M. Neuman, Brian Plancher, and Vijay Janapa Reddi are with the John A. Paulson School of Engineering and Applied Sciences, Harvard University, Cambridge, MA, USA.
{\tt\footnotesize sneuman@seas.harvard.edu, brian\_plancher@g.harvard.edu, vj@eecs.harvard.edu}}%
\thanks{$^{5}$Brian Plancher is with Barnard College, Columbia University, New York, NY, USA.
{\tt\footnotesize bplancher@barnard.edu}}%
}

\begin{document}

\maketitle
\thispagestyle{empty}
\pagestyle{empty}

\begin{abstract}
Hardware acceleration can revolutionize robotics, enabling new applications by speeding up robot response times while remaining power-efficient. However, the diversity of acceleration options makes it difficult for roboticists to easily deploy accelerated systems without expertise in each specific hardware platform.
In this work, we address this challenge with RobotCore, an architecture to integrate hardware acceleration in the widely-used ROS~2 robotics software framework.
This architecture is target-agnostic (supports edge, workstation, data center, or cloud targets) and accelerator-agnostic (supports both FPGAs and GPUs).
It builds on top of the common ROS~2 build system and tools and is easily portable across different research and commercial solutions through a new firmware layer. We also leverage the Linux
Tracing Toolkit next generation (LTTng) to enable low-overhead real-time tracing and benchmarking of accelerated ROS 2 systems.
To demonstrate the acceleration enabled by this architecture, we use it to deploy a ROS~2 perception computational graph on a CPU and FPGA.

We also employ our integrated tracing and benchmarking to analyze bottlenecks, uncovering insights that guide us to improve FPGA communication efficiency. In particular, we design an intra-FPGA ROS 2 node communication queue template and use it in conjunction with FPGA-accelerated nodes to achieve a $24.42\%$ speedup over a CPU.

\end{abstract}

\section{Introduction}
\label{sec:intro}
Recent work has seen an explosion of specialized robotics acceleration on nontraditional computing platforms such as GPUs, FPGAs, and ASICs~\cite{wan2021survey,mayoral2021adaptivecomputing,mayoral2021kria,murray2016robot,murray2016microarchitecture,murray2019programmable,plancher2021accelerating,neuman2021robomorphic,plancher2021grid,austin2020titan,freeman2021brax,suleiman2019navion,liu2020hardware,asgari2020pisces,liu2021archytas,8046383,williams2017model,sacks2018robox,PlancherParallelDDP,PlancherRealtimeMPC,gupta2021efficient}.
This has been sparked by the decline of Moore's Law and Dennard Scaling, which limits the performance of traditional CPU computing, positioning hardware acceleration as an emerging solution to achieve high performance and power efficiency in robotics applications.

However, this increased diversity of computing platforms leads to a dramatic growth in design space complexity that makes it difficult for users to easily deploy robotics applications on hardware accelerators without substantial expertise in each specific accelerator platform.
The Open Computing Language (OpenCL)~\cite{munshi2009opencl} is an effort to standardize hardware acceleration under a common language, but its adoption across silicon vendors has been uneven and support for it varies.
As a result, current hardware acceleration usage is often tied to a particular vendor's solutions and platforms.
This not only impedes interoperability and reuse of acceleration kernels, but presents yet another layer of complexity that users must overcome while implementing robotic systems that use acceleration kernels.
A key obstacle is that each hardware acceleration vendor provides their own framework for development, but these are often disconnected from the common tools and libraries in robotics, and mostly aimed at hardware engineers, not roboticists.

\begin{figure}[!t]
    \centering
    \includegraphics[width=0.85\columnwidth]{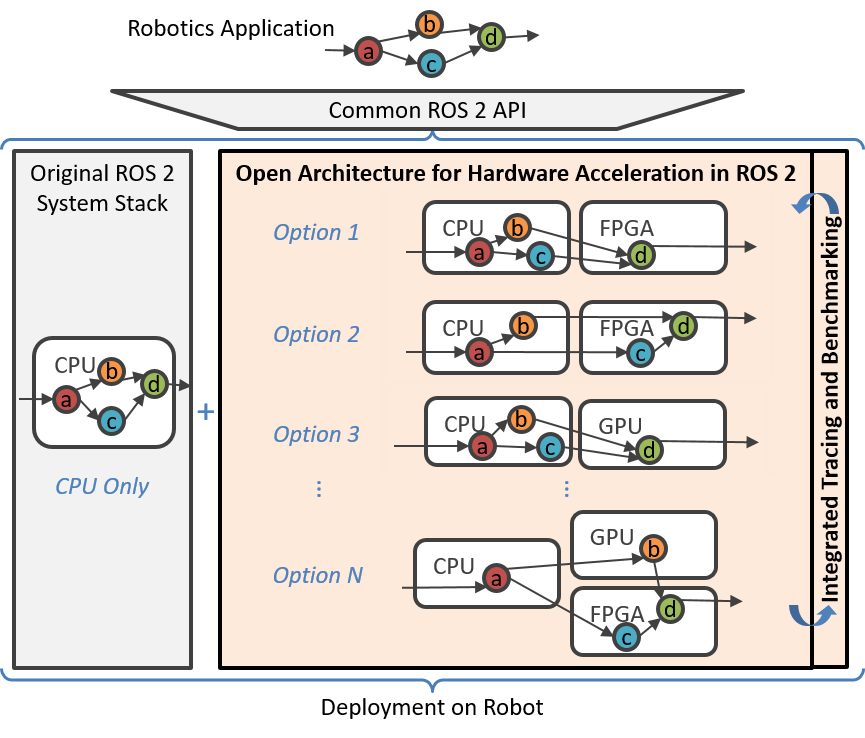}
    \vspace{-10pt}
    \caption{The open architecture for hardware acceleration in ROS~2 extends the ROS~2 build system to support vendor and platform-agnostic deployment of robotics applications on accelerator hardware. The integrated tracing and benchmarking infrastructure enables users to analyze the system and make strategic design improvements to optimize performance.}
    \label{fig:overview}
    \vspace{-20pt}
\end{figure}

To address this challenge,
we present \emph{RobotCore}, an \emph{open architecture for hardware acceleration} that extends the Robot Operating System (ROS)~\cite{quigley2009ros}, the \emph{de facto} standard for robot application development.
ROS is widely used by academia and industry, and early work has demonstrated its potential for hardware-accelerated robotics applications~\cite{mayoral2021adaptivecomputing,mayoral2021kria,nvidia2022isaacros}.
We facilitate this emerging direction by implementing a vendor and platform-agnostic abstraction layer for hardware acceleration in robotics (Fig.~\ref{fig:overview}).
Starting with a popular robotics API as the foundation, our ROS~2-based acceleration architecture provides a common ground for both academic researchers and silicon vendors alike to develop specialized robotics acceleration kernels, and deploy them for easy usage by a large, established user base.

Once roboticists can easily harness hardware acceleration across multiple platforms, the next major challenge is \textit{profiling and benchmarking the application}. Benchmarking is needed to determine the best mapping of the robotics computational graph to the different hardware resources available to optimize overall robot system performance.
This is a difficult task, however, since every application is different and deployment scenarios are widespread. Full end-to-end system analysis is required to understand how different implementation tradeoffs impact overall performance. To enable this analysis, we demonstrate how to leverage prior work~\cite{Bedard_2022} to benchmark accelerated ROS~2 kernels with a low-overhead framework for real-time tracing based on the Linux Tracing Toolkit next generation (LTTng)~\cite{desnoyers2006lttng}. We demonstrate analysis of a case study deployment using CPU and FPGA nodes for a simple perception pipeline. 

Using our framework and benchmarking, we diagnose that substantial latency bottlenecks in this computational graph come from inter-node interactions across ROS~2 layers in the CPU. We recognize this as an opportunity for design optimization in hardware accelerators, because interaction with the CPU should not be necessary for dataflow between nodes co-located on the same non-CPU platform (e.g., FPGA). 

Based on the benchmarking analysis, we demonstrate two novel separate paths toward \emph{hardware acceleration}: (1) kernel fusion, and (2) improved message passing. Kernel fusion results in the highest speedup, an average of $26.96\%$, but it requires manual redesign of the underlying kernels. To avoid manual redesign entirely and improve design re-use and portability, we alternatively develop an intra-FPGA ROS~2 node communication queue template that leverages AXI4-Stream interfaces~\cite{amba4axi4} and transfers data in a sequential streaming manner directly between acceleration kernels. Using this design pattern improves the overall inter-node performance in our computational graph by $24.42~\%$ on average, while requiring no change in the accelerated kernels.
This template extends to applications beyond our case study, since it can be reused for any ROS~2 inter- or intra-process communication by adapting its data types.


In summary, key contributions of this work are that we:
\begin{itemize}
    \item Create a new open infrastructure to increase the performance of robotics applications by enabling \emph{integration of hardware acceleration} into ROS~2 that is \emph{flexible across accelerator platforms} (e.g., FPGAs, GPUs) and system deployments (e.g., edge devices, workstations, data centers, and cloud);
    \item Expose insights into how to optimize overall system-wide performance by extending and providing a template API for \emph{low-overhead tracing and benchmarking framework} to analyze application performance across hardware accelerated ROS~2 computational graphs, laying foundation to analyze mixed-platform systems (e.g., combinations of CPU and FPGA-based nodes); and
    \item Increase ROS~2 node-to-node
    dataflow performance to achieve an average overall accelerator speedup of $24.42\%$ over CPU in our experiments by designing a template for \emph{intra-FPGA ROS~2 node communication queues}, based on insights uncovered using our open acceleration infrastructure and low-overhead benchmarking on a case study of a simple perception graph.
\end{itemize}

The core components of our architecture are disclosed under a commercially friendly open-source license and are available and maintained at the ROS~2 Hardware Acceleration Working Group GitHub organization: \url{https://github.com/ros-acceleration}.

\section{Background and Related Work}
\label{sec:background}
\subsection{ROS and ROS~2}

The Robot Operating System (ROS) is an open-source collection of software frameworks and tools designed to provide a \emph{structured communications layer} for robotics applications running on heterogenous computer hardware~\cite{quigley2009ros}. 

ROS applications are designed around event driven graphs of \emph{Nodes} which communicate through \emph{Messages} on various \emph{Topics}, \emph{Services}, and \emph{Actions}. Each Node can be thought of as a software process which applies an algorithm to the input message and then broadcasts the resulting output message. By managing all inter-Node communications across abstraction layers (e.g., \texttt{rclcpp}, \texttt{rcl}, \texttt{rmw}), ROS simplifies the robotic system deployment process and enables roboticists to quickly develop and test new algorithms.
ROS also provides substantial infrastructure to facilitate the automatic building, evaluation, and deployment of robotic systems, including dependency managers, package managers, build systems and tools, simulators, and visualizers.

ROS~2 is a re-design of ROS that modernizes and updates all of its components while adhering to its core design principles. ROS~2 provides a stronger partitioning of the communication middleware from the robotics logic, enabling more flexibility, scalability, and reliability~\cite{mayoral2021adaptivecomputing}.
ROS~2 also provides an updated build system, \texttt{ament}, and a new universal build tool, \texttt{colcon}. This provides a single simple interface for managing the building and deployment of complete robotics applications. Leveraging these tools, roboticists can write new algorithms and rely on ROS~2 to handle all lower level operations and middleware management.

\subsection{Hardware Acceleration for ROS and ROS~2}

There has been previous work that has focused on ways to accelerate robotics applications by developing tools and methodologies to help roboticists leverage hardware acceleration for select ROS Nodes and to optimize the ROS computational graph through adaptive computing~\cite{yamashina2015proposal,yamashina2016crecomp,podlubne2019fpga,eisoldt2021reconfros,lienen2020reconros,9415584,ohkawa2016architecture,panadda2021low,8956928,9397897,ohkawa2018fpga,9355892,amano2021dataset,nitta2018study,chen2021fogros,nvidia2022isaacros}. There has also been some work to accelerate the scheduling and communication layers used by ROS and ROS~2~\cite{sugata2017acceleration,ohkawa2019high,choi2021picas,suzuki2018real,gutierrez2018time,gutierrez2018real,gutierrez2018towards,gutierrez2018synch}. Unfortunately, the majority of these efforts assume an end-user has substantial experience with embedded systems and embedded hardware flows, or is customized to a specific hardware acceleration board or deployment scenario.

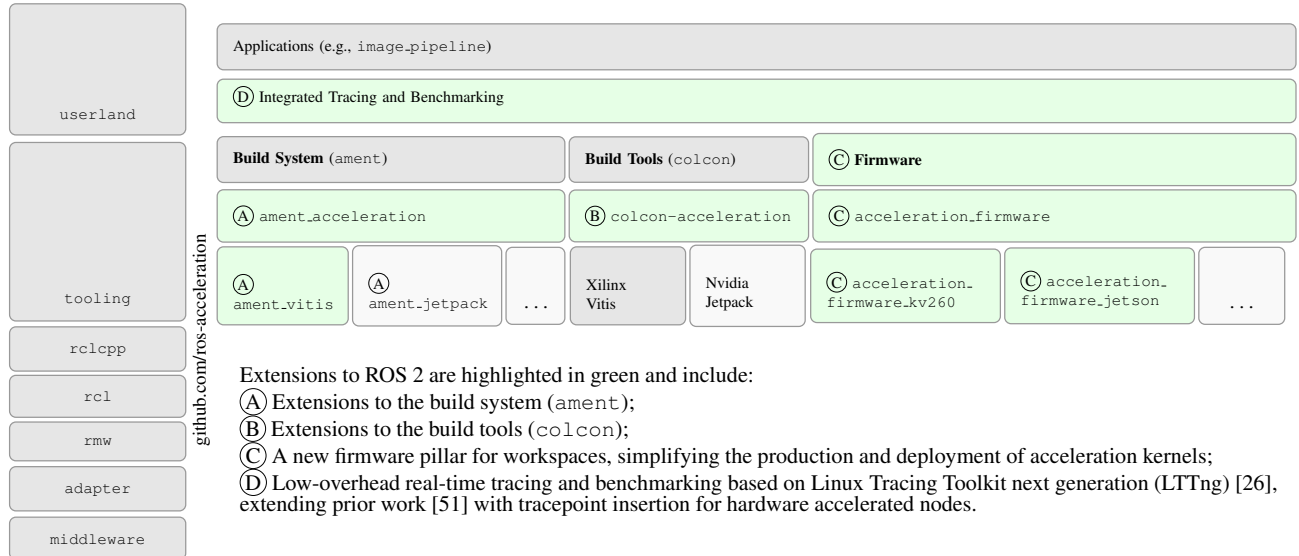
\begin{figure*}[!t]
    \centering
    \scalebox{0.6}{
    \tikzset{
      state/.style={
        rectangle, rounded corners, draw=gray!80, fill=gray!20, thick,
        minimum height=2em, inner sep=10pt, text centered},
      state2/.style={
        rectangle, rounded corners, draw=black, fill=gray!70, thick,
        minimum height=2em, inner sep=10pt, text centered},
      epath/.style={draw, <->, shorten <=1pt, shorten >=1pt},
      cir/.style={draw, circle, fill, inner sep=2.5pt, text=white}
    }
    \begin{tikzpicture}[->, >=latex, line width=0.75pt]
      \draw[fill=white, color=white] (-10.8,-4.5) rectangle (9.2,4.5);
      
      \node[text width=4cm] (mug) at (-12.9,9.) {\Large \tt ROS 2 Stack};
      \node[text width=15cm] (mug) at (2.5,9.) {\Large \tt Open Architecture for Hardware Acceleration};
      
      
      \node[rotate=90] (site) at (-10.6,1) {
        \large github.com/ros-acceleration 
      };
      

    \node[state, text width=7cm, text ragged]
        (buildsystem) at (-6.4, 5.) {
            {\textbf{Build System} (\texttt{ament})}\\
    };
    \node[state, text width=7cm, text ragged, fill=green!10]
        (amentacceleration) at (-6.4, 3.75) {
            {\circledwhite{A} \tt ament\_acceleration}\\
    };

    \node[state, text width=2.2cm, text height=0.6cm, text ragged, fill=green!10]
        (amentvitis) at (-8.8, 2.2) {
            {\circledwhite{A} \tt ament\_vitis}\\
    };

    \node[state, text width=2.6cm, text height=0.6cm, fill=gray!5, text ragged]
        (amentrocm) at (-5.6, 2.2) {
            {\circledwhite{A} \tt ament\_jetpack}\\
    };

    \node[state, text width=0.6cm, text height=1cm, fill=gray!5]
        (amentmore) at (-3.2, 2.2) {
            {\tt ...}\\
    };
    
    \node[state, text width=4.6cm, text ragged]
        (colcon) at (0.2, 5.) {
        {\textbf{Build Tools} (\texttt{colcon})}\\
    };

    \node[state, text width=4.6cm, text ragged, fill=green!10]
        (colconaccel) at (0.2, 3.75) {
        {\circledwhite{B} \tt colcon-acceleration}\\
    };
    
    \node[state, text width=1.85cm, text height=0.6cm, text ragged] (vitis) at (-1.15, 2.2) {
        Xilinx \\Vitis
    };

    \node[state, text width=1.85cm, text height=0.6cm, text ragged, fill=gray!5] (cuda) at (1.5, 2.2) {
        Nvidia\\ Jetpack
    };

    \node[state, text width=10cm, text ragged, fill=green!10]
    (firmware) at (8.3, 5.) {
        {\circledwhite{C} \textbf{Firmware}}\\
    };
    \node[state, text width=10cm, text ragged, fill=green!10]
    (accelerationfirmware) at (8.3, 3.75) {
        {\circledwhite{C} \tt acceleration\_firmware}\\
    };
    \node[state, text width=3.5cm, text height=0.5cm, text ragged, fill=green!10]
        (accelerationfirmwarekv260) at (5.0, 2.2) {
            {\circledwhite{C} \tt acceleration\_\\firmware\_kv260}
    };
    \node[state, text width=3.5cm, text height=0.5cm, text ragged, fill=green!10]
        (accelerationfirmwarejetson) at (9.3, 2.2) {
            {\circledwhite{C} \tt acceleration\_\\firmware\_jetson}
    };
    \node[state, text width=1.2cm, text height=1cm, fill=gray!5]
        (accelerationfirmwaremore) at (12.45, 2.2) {
            {\tt ...}
    };

    \node[state, text height=0.15cm, text width=23.2cm, text ragged, fill=green!10]
        (tracing) at (1.7, 6.3) {
            {\circledwhite{D} Integrated Tracing and Benchmarking}\\
    };
    \node[state, text width=23.2cm, text ragged]
        (accelerationexamples) at (1.7, 7.5) {
            {Applications (e.g., {\tt image\_pipeline})}\\
    };

    \node[state, text width=3.2cm, text height=2.2cm]
      (userland) at (-12.9, 7.) {
        {\tt userland}\\
    };
    \node[state, text width=3.2cm, text height=3.2cm]
      (tooling) at (-12.9, 3.4) {
        {\tt tooling}\\
    };

    \node[state, text width=3.2cm]
      (rcl) at (-12.9, 0.8) {
        {\tt rclcpp}\\
    };

    \node[state, text width=3.2cm]
      (rcl) at (-12.9, -0.25) {
        {\tt rcl}\\
    };
    \node[state, text width=3.2cm]
      (rmw) at (-12.9,-1.25) {
        {\tt rmw}\\
    };
    \node[state, text width=3.2cm]
      (adapter) at (-12.9,-2.3) {
        {\tt adapter}\\
    };
    \node[state, text width=3.2cm]
      (middleware) at (-12.9,-3.4) {
        {\tt middleware}\\
    };

    \node[text width=23.5cm] (pilar1) at (2,-1.25) {\Large Extensions to ROS 2 are highlighted in green and include:\\\circledwhite{A} Extensions to the build system (\texttt{ament}); \\\circledwhite{B} Extensions to the build tools (\texttt{colcon}); \\\circledwhite{C} A new firmware pillar for workspaces, simplifying the production and deployment of acceleration kernels; \\\circledwhite{D} Low-overhead real-time tracing and benchmarking based on Linux Tracing Toolkit next generation (LTTng)~\cite{desnoyers2006lttng}, extending prior work~\cite{bedard2022ros2_tracing} with tracepoint insertion for hardware accelerated nodes.};

    \end{tikzpicture}}
    \vspace{-20pt}
    \caption{Overview of the components of the open architecture for hardware acceleration in ROS~2.}
    \label{fig:architecture}
    \vspace{-10pt}
\end{figure*}

Our proposed open architecture takes a ROS-centric approach to integrate the hardware and embedded flows directly into the core ROS~2 ecosystem. This enables a separation between those who produce accelerated kernels and those who use them by providing end-users with a build and deployment experience for hardware accelerators similar to the standard, non-accelerated ROS~2 experience.

\section{An Open Architecture for Hardware Acceleration in ROS~2}
\label{sec:architecture}

Our open architecture (Fig.~\ref{fig:architecture}) extends the core ROS~2 build system and tools to provide platform-agnostic (i.e., supports  edge, workstation, data center, or cloud targets) and technology-agnostic (i.e., supports FPGAs and GPUs), hardware-accelerated ROS~2 capabilities for roboticists. We: A)~extend the ROS~2 build system, \texttt{ament}; B)~extend the ROS~2 meta build tool, \texttt{colcon}; and C)~develop integrated ROS~2 firmware extensions. We also D)~integrate a low-overhead tracing and benchmarking framework to enable the analysis of holistic application performance across ROS graphs. This section describes these extensions in detail.

\pgfkeys{/csteps/inner color=white}
\pgfkeys{/csteps/fill color=black}

\subsection{Extending the ROS~2 Build System}
\label{sec:architecture:step1}

The first pillar of our open architecture, Fig.~\ref{fig:architecture}~\circledwhite{A}, allows roboticists to generate acceleration kernels directly from the ROS~2 build system (\texttt{ament}) in the same way they generate CPU binaries. To do so, the \texttt{ament\_acceleration} ROS~2 package and its extensions abstract the ROS build system from vendor-specific accelerators (e.g. FPGAs or GPUs), including their frameworks and software platforms. This allows the build system to easily support hardware acceleration across commercial solutions while using the same syntax, simplifying the work of ROS~2 package maintainers. 

Under the hood, each hardware-specific extension of \texttt{ament\_acceleration} abstracts away the corresponding vendor-specific firmware. For example, \texttt{ament\_vitis}\footnote{{github.com/ros-acceleration/ament\_vitis}} relies on the proprietary Xilinx Vitis~\cite{xilinx2022vitis} and on the Xilinx Runtime (XRT) library~\cite{xilinx2022xrt}. 
This simplifies the creation of acceleration kernels and separates firmware concerns from algorithm development. This way, robotics engineers can focus on improving their computational graphs with a ROS-centric development flow. Separately, hardware experts, potentially sponsored by silicon vendors, can improve acceleration kernels for a particular commercial solution. Overall, these extensions help achieve the objective of simplifying the creation and integration of acceleration kernels from different vendors into ROS~2 computational graphs.


Fig.~\ref{fig:architecture} depicts the build system extensions showing how  \texttt{ament\_acceleration} abstracts the build system from vendor-specific solutions. As an example of an alternative acceleration technology supported, \texttt{ament\_jetpack} is included and illustrates the integration of Nvidia JetPack~\cite{nvidia2022jetpack}. 


\subsection{Extending the ROS~2 Build Tools}
\label{sec:architecture:step2}

The second pillar of our open architecture, Fig.~\ref{fig:architecture}~\circledwhite{B}, extends the \texttt{colcon} ROS~2 meta build tool to integrate hardware acceleration flows into the ROS~2 Command Line Interface (CLI) commands. Examples of these extensions include the selection of the target accelerator and build-time through mixins, emulation capabilities to speed-up the development process and facilitate design without access to the real hardware, raw disk image production tools, and simplified configuration of hypervisors. These extensions are implemented by the \texttt{colcon-acceleration}\footnote{{github.com/ros-acceleration/colcon-acceleration}} ROS~2 package. As in Section~\ref{sec:architecture:step1}, \texttt{colcon\_acceleration} further enables roboticists to leverage hardware accelerators while using standard ROS~2 commands and flows.

\subsection{ Adding Firmware Extensions}
\label{sec:architecture:step3}

Represented by the abstract \texttt{acceleration\_firmware} ROS package and its corresponding specializations (e.g. \texttt{acceleration\_firmware\_kv260}\footnote{{github.com/ros-acceleration/acceleration\_firmware\_kv260}} for the Xilinx Kria KV260 board), the third pillar of our open architecture, Fig.~\ref{fig:architecture}~\circledwhite{C}, firmware extensions, are meant to provide firmware artifacts for each supported technology solution. This again simplifies the process for ROS package consumers and maintainers, and further aligns hardware acceleration workflows with typical ROS development flows.
Each ROS~2 workspace can leverage multiple firmware packages, but can only use one at a time. As \texttt{colcon\_acceleration} supports the selection of the active firmware in the ROS workspace, by separating the firmware out into their own packages, our open architecture enables silicon vendors to maintain an \texttt{acceleration\_firmware\_<solution>} package that automatically integrates into standard ROS~2 workflows. 

\subsection{Low-Overhead Real-Time Tracing \& Benchmarking}
\label{sec:architecture:step4}

In the context of hardware acceleration in robotics, it is fundamental to be able to inspect performance improvements. To that end, it is important to benchmark and trace the system. Benchmarking is the process of running a computer program to assess its relative performance, whereas tracing is a technique used to understand what is happening in a system while it is running. Tracing helps determine which pieces of a Node are consuming more compute cycles or generating indeterminism, and are thereby good candidates for hardware acceleration. Benchmarking instead helps investigate the relative performance of an acceleration kernel versus its CPU scalar computing baseline. Similarly, benchmarking also helps with comparing acceleration kernels across different hardware acceleration technology solutions (e.g., Kria KV260 vs. Jetson Nano) and across kernel implementations within the same hardware acceleration technology solution.

\begin{figure}[!t]
    \centering
    \includegraphics[width=1.0\columnwidth]{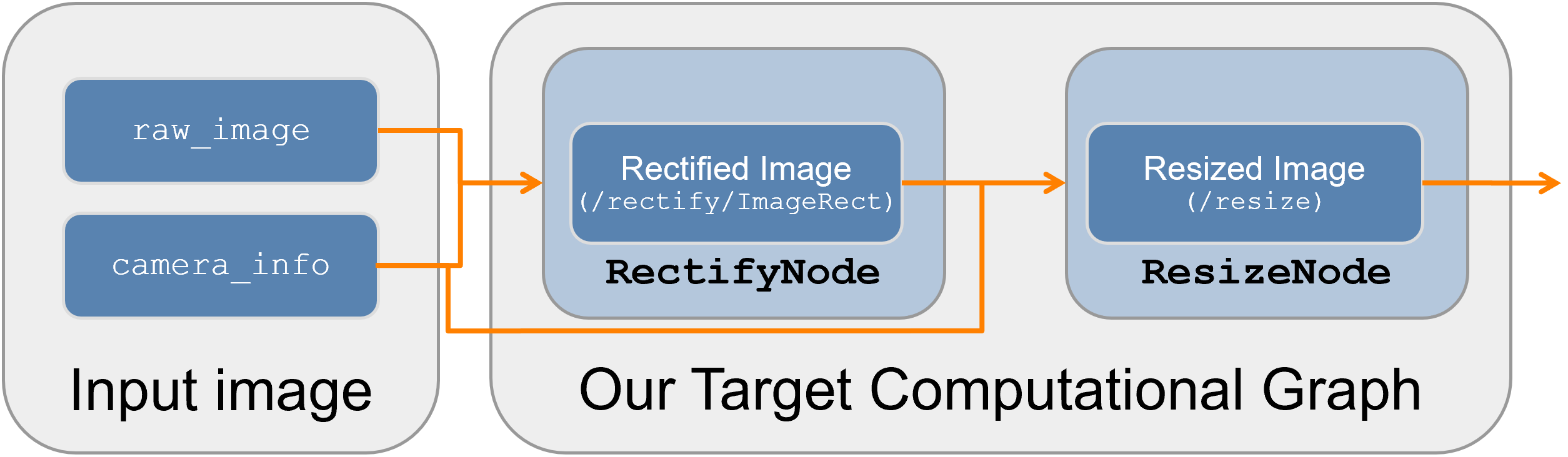}
    \vspace{-15pt}
    \caption{Computational graph of our case study perception application, \texttt{image\_pipeline}, containing two ROS~2 nodes: (1)~\texttt{RectifyNode} subscribes to the \texttt{/camera/image\_raw} and \texttt{/camera/camera\_info} topics from Gazebo \cite{koenig2004design} and publishes a \emph{rectified image} to (2)~\texttt{ResizeNode}, which publishes the final \emph{resized image}.}
    \label{fig:usecase1}
\end{figure}

In order to trace and evaluate the relative performance of both ROS~2 individual Nodes and complete computational graphs, we leverage Linux Tracing Toolkit next generation (LTTng \cite{desnoyers2006lttng}) for tracing and benchmarking, Fig.~\ref{fig:architecture}~\circledwhite{D}. Building upon prior work \cite{bedard2022ros2_tracing}, LTTng provides a collection of flexible tracing tools and multipurpose instrumentation for ROS~2 that allow collecting runtime execution information in real-time in distributed systems using low-overhead tracers. 
For example, when enabling all ROS~2 instrumentation, end-to-end message latency overhead is below $5.5$us \cite{bedard2022ros2_tracing}, making it suitable for a wide variety of hardware acceleration use cases.
Building on top of this foundation, we developed a tracing and benchmarking template that enables roboticists to easily instrument both their accelerated and non-accelerated code in a vendor-agnostic manner. This infrastructure also lays a foundation for future integration with platform-specific performance counters and tracing tools that can extend analysis to more fine-grained introspection and profiling of the kernels running onboard an accelerator device.



\section{Case Study: Accelerating ROS~2 Perception}
\label{sec:casestudy}
For our case study, we trace, benchmark, and accelerate a subset of  \texttt{image\_pipeline}~\cite{mihelich2022image_pipeline},
one of the most popular packages in the ROS~2 ecosystem, and a core piece of the ROS perception stack. We compose a simple computational graph consisting of two nodes, \texttt{resize} and \texttt{rectify}, as shown in Fig.~\ref{fig:usecase1}. We then leverage our open architecture for hardware acceleration (Section~\ref{sec:architecture}) to benchmark, trace and accelerate our computational graph, comparing a CPU to an FPGA implementation. In this section we describe the methodology of our approach, and analyze our timing results, presenting a case study for how our open architecture can help enable hardware accelerated applications in ROS~2.




\begin{figure*}[!t]
    \centering
    \includegraphics[width=0.91\textwidth]{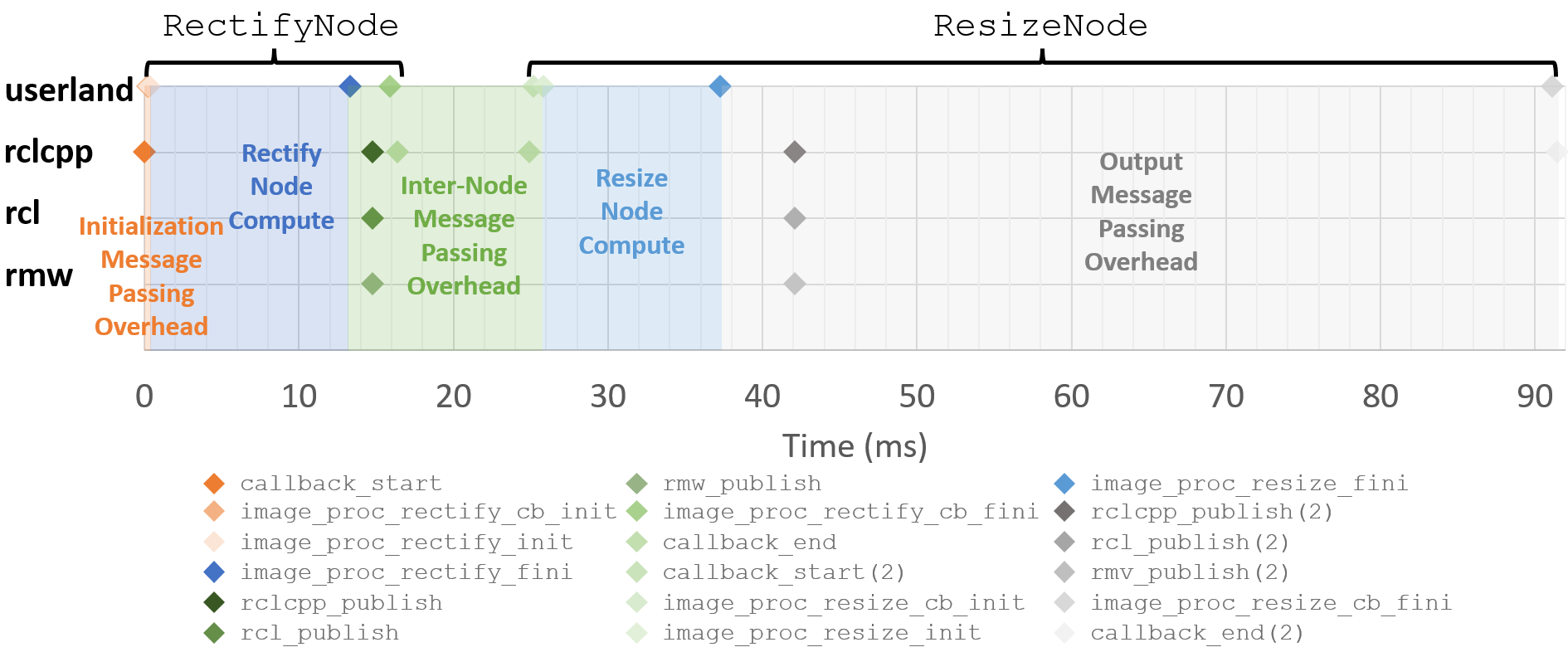}
    \caption{Tracepoints instrumented across ROS~2 abstraction layers on CPU for case study computational graph (Fig.~\ref{fig:usecase1}). Breakdown summary in Fig.~\ref{fig:benchmarkcpufpga2}.}
    \label{fig:instrumentation2}
    \vspace{-10pt}
\end{figure*}

\subsection{Methodology}
\label{sec:casestudy:methodology}
We propose the following steps
to analyze a ROS~2 application and design appropriate acceleration:
(i)~instrument both the core components of ROS~2 and the target kernels; (ii)~trace and benchmark the kernels on the CPU to establish a baseline; (iii)~develop a hardware accelerated implementation on alternate hardware (e.g., GPU, FPGA); and (iv)~trace, benchmark against the CPU baseline, and improve the accelerated implementation.

Following this methodology, in our case study we begin by instrumenting both ROS~2 and our target kernels with LTTng probes. Reusing past work and probes~\cite{bedard2022ros2_tracing} allows us to easily get a grasp of the dataflow interactions within \texttt{rmw}, \texttt{rcl}, and \texttt{rclcpp} ROS~2 layers. We then also instrument the \texttt{ResizeNode} and \texttt{RectifyNode} components of the \texttt{image\_pipeline} package used in our case study.
The relevant tracepoints placed in our computational graph across ROS~2 stack layers are listed in Fig.~\ref{fig:instrumentation2}~and~\ref{fig:benchmarkcpufpga2} (full list in Pull Request $717$ in the \texttt{image\_pipeline} repository~\cite{mihelich2022image_pipeline}).
On the CPU, these tracepoints enable us to isolate the latency of computation within a node from the time it takes ROS~2 to package and pass information between nodes.


In the following sections we report timing results from using a Xilinx Kria® KV260 Vision AI Starter Kit~\cite{xilinx2022kria}, which has an onboard integrated Quad-core Arm® Cortex®-A53 CPU and an FPGA containing 256K System Logic Cells and 1.2K DSP Slices. All benchmark results report the mean value obtained from a $60$ second continuous run of the computational graph. The FPGA kernels are synthesized, placed and routed with a $250$MHz clock.

\subsection{CPU-Only Tracing Results}
\label{sec:casestudy:cpuresults}

\begin{figure}[!t]
    \centering
    \includegraphics[trim=0 0 0 12, clip, width=1.0\columnwidth]{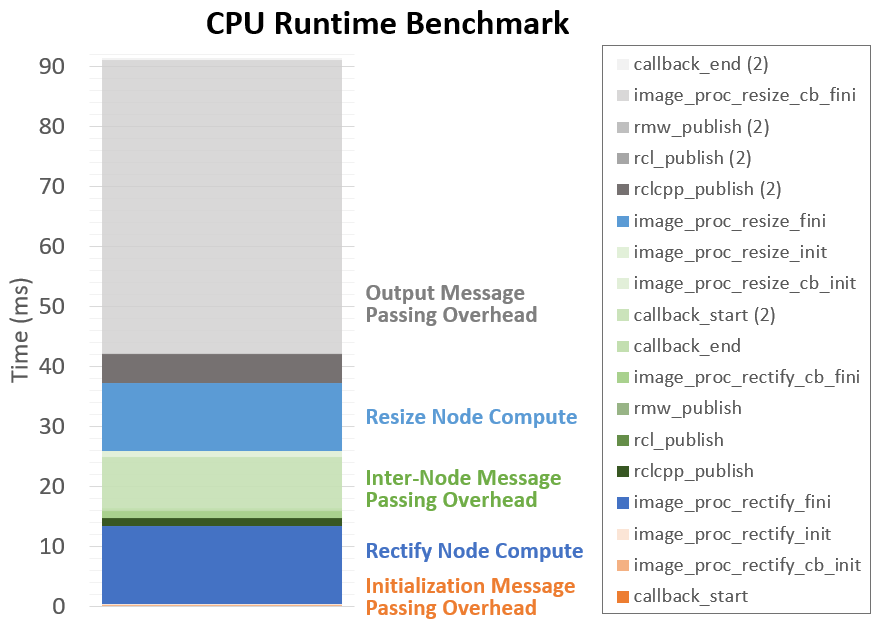}
    \vspace{-20pt}
    \caption{Breakdown of CPU runtime derived from tracing and benchmarking. Total computation time of our case study graph is dominated by message passing overheads, a bottleneck consuming over $73.3\%$ of total runtime.}
    \label{fig:benchmarkcpufpga2}
\end{figure}

Fig.~\ref{fig:instrumentation2} demonstrates the results of instrumenting and tracing our target computational graph (Fig.~\ref{fig:usecase1}) across multiple ROS~2 stack layers on the CPU,
and Fig.~\ref{fig:benchmarkcpufpga2} summarizes the breakdown of timing results across operations, establishing the CPU baseline for our application.
The breakdown in Fig.~\ref{fig:benchmarkcpufpga2} shows the time taken to do the computations within each node, as well as the time taken by the ROS~2 lower-level message-passing system across the various abstraction layers. We find that the message-passing overhead in our application consumes more than $73.3\%$ of the total time and is therefore a large bottleneck in the total computation time of the full graph.
We next explore FPGA hardware acceleration options, comparing performance to the CPU baseline.


\subsection{Accelerating and Benchmarking CPU \& FPGA}
\label{sec:casestudy:acceleration}
In this section, we explore hardware acceleration options for an FPGA for our case study application (Fig.~\ref{fig:usecase1}).
In Section~\ref{sec:casestudy:acceleration:nodes}, we first explore hardware acceleration kernels for the core logic of each of the Nodes (\emph{rectify} and \emph{resize}), harnessing our open architecture for implementation.
In Section~\ref{sec:casestudy:acceleration:graph}, we then explore two different FPGA designs to accelerate the computational graph by optimizing dataflow interactions between FPGA-based nodes, addressing the ROS~2 \emph{communication infrastructure} performance bottleneck revealed by the CPU baseline in Section~\ref{sec:casestudy:cpuresults}.

\subsubsection{Accelerating \underline{Nodes \& Components} on an FPGA}
\label{sec:casestudy:acceleration:nodes}
We first accelerate the computations at each one of the graph nodes. The \texttt{RectifyNode} and \texttt{ResizeNode} \emph{Components} of Fig.~\ref{fig:usecase1} are accelerated using Xilinx's HLS, XRT, and OpenCL targeting the Kria KV260\footnote{github.com/ros-acceleration/image\_pipeline/blob/ros2/image\_proc/src/ \{rectify,resize\}\_fpga.cpp}.
Each ROS~2 \emph{Component} has an associated acceleration kernel\footnote{{github.com/ros-acceleration/image\_pipeline/tree/ros2/image\_proc/src/ image\_proc}} that leverages the Vitis Vision Library, a computer vision library optimized for Xilinx silicon solutions and based on OpenCV APIs. These accelerated \emph{Components} and their kernels easily integrate with the rest of the ROS meta-package through our open architecture (Fig.~\ref{fig:architecture}), and are openly available to the public. Building the accelerators is abstracted away from roboticist end-users, and takes no significant additional effort than the standard build of the \texttt{image\_pipeline}.

After benchmarking the accelerated \emph{Components} using the trace points of Section~\ref{sec:casestudy:cpuresults}, we observe an average $6.22\%$ speedup in the total computation time of the perception pipeline when offloading tasks to the FPGA (see Fig.~\ref{fig:benchmarkintegrated2}).
For this case study example, it is not surprising that accelerating the computational nodes and components alone only gives a modest performance increase because, as we saw in Section~\ref{sec:casestudy:cpuresults}, the performance bottleneck in the baseline CPU system was communication overhead, not computation.

\subsubsection{Accelerating the \underline{Computational Graph} on an FPGA}
\label{sec:casestudy:acceleration:graph}
In our case study application, message-passing overheads across the ROS~2 abstraction layers far outweigh other operations, so in this section we focus on optimizing these dataflows.
Addressing performance bottlenecks in our system leads to overall lower computational graph latency, and to faster robots. 
To seize this acceleration opportunity in our case study example, we optimize the dataflow within the computational graph and across ROS~2 Nodes and Components through two different design approaches: (a) kernel fusion, and (b) dedicated streaming queues. 

\begin{figure}[!t]
    \centering
    \includegraphics[width=0.93\columnwidth]{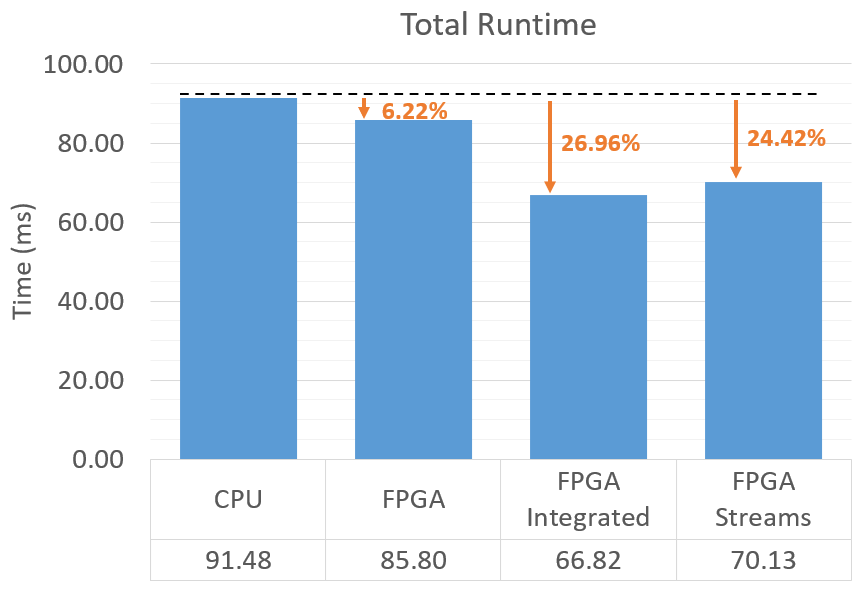}
    \caption{Total runtime of CPU baseline and FPGA, FPGA-Integrated, and FPGA-Streaming hardware-accelerated implementations of case study application.
    Acceleration enables up to $26.96\%$ speedup over CPU.}
    \label{fig:benchmarkintegrated2}
\end{figure}

The speedup obtained by integrating both ROS \emph{Components} on the FPGA into a \emph{single unified kernel} is shown in Fig.~\ref{fig:benchmarkintegrated2}. The benefits of doing this are two-fold. First, we avoid any message-passing between the \emph{Rectify} and \emph{Resize} Nodes' \emph{Components}. Second, we avoid the compute cycles wasted while memory is mapped back and forth between the host CPU and the FPGA. This results in an overall latency speedup of $26.96\%$ over the CPU. In addition to speeding up the perception stage, another added benefit of this improvement is that such speedups make room for other robot tasks in a complete end-to-end system. Note, however, that this improvement required the construction of an entirely new ROS Node and unified acceleration kernel on the FPGA.

We then develop a template for an \emph{accelerated ROS~2 message passing} interface on the FPGA. This interface is \emph{Node} and \emph{Component}-agnostic and can be leveraged by roboticists to accelerate the communication channels of any computational graph on an FPGA. 
This is done by leveraging an AXI4-Stream interface to create an intra-FPGA ROS~2 communication queue template which is then used to pass data across Nodes in the FPGA without sending messages to the CPU\footnote{AXI4-Stream interfaces are data-type specific and as such our template may require type adaptations for other use cases depending on the Node-to-Node data interactions.}. This allows us to completely bypass the original CPU-centric ROS~2 message-passing system and optimizes dataflow, achieving an overall latency improvement of $24.42\%$ over the CPU in our application (see Fig.~\ref{fig:benchmarkintegrated2}).


Based on these results, for this case study, we show that implementing FPGA-accelerated versions of key ROS~2 \emph{Components} is easily feasible, and that addressing the right bottleneck is key to improving performance.
Tracing and benchmarking the CPU baseline suggested that communication is the bottleneck in our case study.
In fact, independent examination of, e.g., a single run of the fused-kernel accelerator using the Xilinx Vitis Analyzer, confirms that this is also the case on the FPGA---we note that integrating device-specific profiling tools into our foundational tracing infrastructure in future work can further automate this type of fine-grained introspection of kernels onboard accelerator devices.
We can achieve overall performance improvements by either combining Nodes or streamlining intra-FPGA communication.
While combining nodes may result in slightly higher performance, it is a much more labor-intensive design effort. By contrast, our accelerated intra-FPGA-Node communication queue template can be applied by any roboticist, to any computational graph.
\section{Conclusion and Future Work}
\label{sec:conclusion}

In this work we present a new open infrastructure to introduce hardware acceleration in ROS~2 in a scalable and technology-agnostic manner.
Our architecture allows us to increase the performance of robotics applications through the \emph{integration of hardware acceleration} with ROS~2 APIs and its conventional flows.
We do so by extending ROS~2 in a way that is \emph{portable across accelerator platforms} (e.g., FPGAs, GPUs) and system deployments (e.g., edge devices, workstations, data centers, and cloud).
We also present a \emph{template for low-overhead tracing and benchmarking} to analyze performance across both hardware accelerated and standard ROS~2 computational graphs.

We use our open architecture and our tracing and benchmarking infrastructure to demonstrate a principled design methodology for ROS~2 hardware acceleration, exposing insights into how to optimize overall system-wide performance by analyzing a CPU baseline, and comparing accelerator design iterations to that original baseline.
We examine a case study using the Xilinx Kria KV260 platform to demonstrate FPGA acceleration of one of the most popular packages in the ROS perception pipeline: \texttt{image\_pipeline}.
We first demonstrate a modest performance speedup of $6.22\%$ from offloading perception tasks to the FPGA,
and then increased speedup by additionally addressing the communication overheads that we identified as bottlenecks by analyzing our CPU baseline.
We achieved a speedup of $26.96\%$ from re-architecting the graph to combine nodes and avoid inter-FPGA-node communication delays inflicted by interactions with the CPU, but this approach requires substantial effort from users to re-architect their graphs.
Instead, to avoid this overhead and stay in alignment with the ROS~2 programming model, we then design a novel \emph{template for intra-FPGA ROS~2 Node communication queues} that allows ROS \emph{Nodes} and \emph{Components} to deliver faster dataflows, achieving a $24.42\%$ speedup over a CPU without excessive manual per-kernel design effort.

We contribute our open architecture to the ROS community, so that future work can use our infrastructure and extend to new applications beyond our case study example. Promising directions for future work include: benchmarking computational graphs with other hardware solutions (e.g., GPUs) to establish consistent cross-accelerator comparisons; extending our tracing and benchmarking approach to include additional tracing information (e.g., profiling within FPGA or GPU devices) for more fine-grained introspection of kernels running onboard accelerators; and applying our open architecture and analysis to other ROS~2 packages.

Our code is disclosed under a commercially friendly open-source license and is available and maintained at the ROS~2 Hardware Acceleration Working Group GitHub organization: \url{https://github.com/ros-acceleration}.
This work is being further integrated into the ROS ecosystem through a community standardization effort, REP-2008~\cite{hawg2022rep2008}.

\bibliographystyle{bib/IEEEtran}
\bibliography{bib/IEEEabrv,bib/bibliography}

\end{document}